\documentclass[11pt]{article}
\usepackage{fullpage,amsmath,amssymb,ifthen,stmaryrd,natbib}
\bibpunct{(}{)}{,}{a}{}{;}

\title{Lambda Dependency-Based Compositional Semantics}
\newcommand\sE{\ensuremath{\mathcal{E}}}

\newcommand\sK{\ensuremath{\mathcal{K}}}

\newcommand\sP{\ensuremath{\mathcal{P}}}

\newcommand\sV{\ensuremath{\mathcal{V}}}

\newcommand\bR{\ensuremath{\mathbf{R}}}

\ifthenelse{\isundefined{\definition}}{\newtheorem{definition}{Definition}}{}
\ifthenelse{\isundefined{\assumption}}{\newtheorem{assumption}{Assumption}}{}
\ifthenelse{\isundefined{\proposition}}{\newtheorem{proposition}{Proposition}}{}
\ifthenelse{\isundefined{\theorem}}{\newtheorem{theorem}{Theorem}}{}
\ifthenelse{\isundefined{\lemma}}{\newtheorem{lemma}{Lemma}}{}
\ifthenelse{\isundefined{\corollary}}{\newtheorem{corollary}{Corollary}}{}
\ifthenelse{\isundefined{\alg}}{\newtheorem{alg}{Algorithm}}{}
\ifthenelse{\isundefined{\example}}{\newtheorem{example}{Example}}{}
\newcommand\nl[1]{{\it``#1''}} % Natural language
\newcommand\wl[1]{\text{\footnotesize{\tt #1}}} % Logical Language (in the world)
\newcommand\tolc[1]{\llbracket #1 \rrbracket} % Denotation with enviroment
\newcommand\reverse[1]{\bR[#1]} % reverse
\author{Percy Liang}

\begin{document}
\maketitle
\begin{abstract}
This short note presents a new formal language, lambda dependency-based compositional semantics (lambda DCS) for representing logical forms in semantic parsing.  By eliminating variables and making existential quantification implicit, lambda DCS logical forms are generally more compact than those in lambda calculus.
\end{abstract}

\section{Introduction}

Semantic parsing is the task of mapping natural language utterances to logical
forms in some formal language such as lambda calculus.
This short note describes \emph{lambda dependency-based compositional
semantics} (lambda DCS), an alternate formal language which can be
notationally simpler than lambda calculus.  Here is an example:\footnote{This example
  is based on the Freebase schema, which reifies events (called compound value types in Freebase).
  Here, $\wl{PlacesLived}(x,e)$ denotes that a person $x$
  was involved in a living \emph{event} $e$, and $\wl{Location}(e, \wl{Seattle})$
  denotes that the location of that event is Seattle.
Other properties of $e$ would include $\wl{StartDate}(e, \cdot)$ and $\wl{EndDate}(e, \cdot)$.}
\begin{itemize}
  \item Utterance: \nl{people who have lived in Seattle}
  \item Logical form (lambda calculus): $\lambda x . \exists e . \wl{PlacesLived}(x, e) \wedge \wl{Location}(e, \wl{Seattle})$
  \item Logical form (lambda DCS): $\wl{PlacesLived}.\wl{Location}.\wl{Seattle}$
\end{itemize}
As one can see, lambda DCS attempts to remove explicit use of variables.
This makes it similar in spirit to dependency-based compositional semantics (DCS) \citep{liang11dcs},
but DCS is restricted to tree-structured logical forms, and is therefore not as expressive as lambda calculus.
Lambda DCS, on the other hand, borrows the use of variables and lambda abstraction from lambda calculus
when necessary to handle phenomena such as anaphora.

Lambda DCS was designed in the context of building a natural language interface
into Freebase \citep{bollacker2008freebase}, a large graph database, where the
logical forms are database queries (see \citet{berant2013freebase} for our companion paper).
Therefore, we will focus on representing the
semantics of noun phrases or wh-questions.  Also, we will discuss only
the semantics of the formal language, not the compositional semantics
of mapping natural language utterances into lambda DCS.
Finally, this document assumes the reader has basic familiarity with lambda
calculus in the context of natural language (see
\citet{carpenter98type} for an introduction).

%%%%%%%%%%%%%%%%%%%%%%%%%%%%%%%%%%%%%%%%%%%%%%%%%%%%%%%%%%%%

\section{Definition}

We now present the full definition of lambda DCS.
In the background, we have a \emph{knowledge base} of assertions.
Let $\sE$ be a set of entities (e.g., $\wl{Seattle} \in \sE$),
$\sP$ be the set of properties (e.g., $\wl{PlaceOfBirth} \in \sE$).
Then $\sK \subset \sE \times \sP \times \sE$ is the knowledge base,
which can be visualized as a directed graph where nodes are entities
and edges represent assertions in the knowledge graph.
Also, let $\sV$ be a set of variables (e.g., $x_1, x_2 \in \sV$).

Let $[\text{condition}]$ denote the truth value of the condition.
For example, $\lambda x . [x = 3]$ denotes the function that returns true
if and only if its argument is equal to $3$.

% General commentary
We will now walk through each of the constructs in lambda DCS with concrete examples,
and also provide a formal definition of their semantics by providing
a formal conversion to equivalent lambda calculus forms.
Notationally, let $\tolc{z}$ be the lambda calculus form corresponding
to the lambda DCS form $z$.
In the following definitions,
let $x,y$ denote fresh variables which are not used anywhere else.

%%%%%%%%%%%%%%%%%%%%%%%%%%%%%%
\paragraph{Unary base case}

The simplest lambda DCS form is a single entity.
For example,
\begin{align}
\wl{Seattle}
\end{align}
denotes the singleton set containing Seattle.
The equivalent in lambda calculus is:
\begin{align}
\lambda x . [x = \wl{Seattle}].
\end{align}

In general, for an entity $e \in \sE$,
$e$ is a unary logical form
representing the indicator function that returns true when its argument is
equal to $e$:
\begin{align}
\tolc{e} = \lambda x . [x = e].
\end{align}

%%%%%%%%%%%%%%%%%%%%%%%%%%%%%%
\paragraph{Binary base case}

The next example is
\begin{align}
\wl{PlaceOfBirth},
\end{align}
which denotes the set of pairs of people and where they were born:
\begin{align}
\lambda x . \lambda y . \wl{PlaceOfBirth}(x, y).
\end{align}
In general, for a property $p \in \sP$,
$p$ is a binary logical form,
which denotes a function mapping two arguments to whether $p$ holds:
\begin{align}
\tolc{p} = \lambda x . \lambda y . p(x, y).
\end{align}

%%%%%%%%%%%%%%%%%%%%%%%%%%%%%%
\paragraph{Join}

The most central operation in lambda DCS is join.
For example, \nl{people born in Seattle} is represented as
\begin{align}
\wl{PlaceOfBirth}.\wl{Seattle}
\end{align}
in lambda DCS and as
\begin{align}
\lambda x . \wl{PlaceOfBirth}(x, \wl{Seattle})
\end{align}
in lambda calculus.
In general, for a binary logical form $b$ and unary logical form $u$,
we have that $b.u$ is a unary logical form
which denotes:
\begin{align}
\tolc{b.u} = \lambda x . \exists y . \tolc{b}(x, y) \wedge \tolc{u}(y).
\end{align}
The key feature of a join is the \emph{implicit existential quantification} over the
argument $y$ shared by $b$ and $u$.  From a database perspective,
this is simply a join over relations $b$ and $u$ on the second argument of $b$
and the first (and only) argument of $u$ and then projecting out the joined
variable.

The advantage of lambda DCS is more apparent when binaries are chained.
For example, consider \nl{those who had children born in Seattle}:
\begin{align}
\wl{Children}.\wl{PlaceOfBirth}.\wl{Seattle},
\end{align}
and its lambda calculus equivalent:
\begin{align}
\lambda x . \exists y . & \wl{Children}(x, y) \wedge \wl{PlaceOfBirth}(y, \wl{Seattle}).
\end{align}
Here, $\wl{Children}(x,y)$ denotes whether $x$ has a child $y$
and $\wl{PlaceOfBirth}(y,\wl{Seattle})$ denotes whether $y$ was born in Seattle.

A logical form $p_1 . \cdots . p_k . e$ corresponds to a chain-structured graph pattern
on the knowledge base $\sK$ which matches any entity (node) which can reach $e$
by following $k$ edges labeled with the given properties $p_1, \dots, p_k$.

%%%%%%%%%%%%%%%%%%%%%%%%%%%%%%
\paragraph{Intersection}

The set of scientists born in Seattle in lambda DCS:
\begin{align}
\wl{Profession}.\wl{Scientist} \sqcap \wl{PlaceOfBirth}.\wl{Seattle}
\end{align}
and in lambda calculus:
\begin{align}
\lambda x . & \wl{Profession}(x, \wl{Scientist}) \wedge \wl{PlaceOfBirth}(x, \wl{Seattle}).
\end{align}
In general, for two unaries $u_1$ and $u_2$
$u_1 \sqcap u_2$ is a unary logical form representing:
\begin{align}
\tolc{u_1 \sqcap u_2} = \lambda x . \tolc{u_1}(x) \wedge \tolc{u_2}(x).
\end{align}

In terms of the graph pattern perspective, intersection allows tree-structured
graph patterns, where branch points correspond to the intersections.

\paragraph{Union}

Whereas intersection corresponds to conjunction, union corresponds to disjunction.
The set containing \nl{Oregon, Washington and Canadian provinces} in lambda DCS:
\begin{align}
\wl{Oregon} \sqcup \wl{Washington} \sqcup \wl{Type}.\wl{CanadianProvince}
\end{align}
and in lambda calculus:
\begin{align}
\lambda x . [x = \wl{Oregon}] \vee [x = \wl{Washington}] \vee
            \wl{Type}(x, \wl{CanadianProvince}).
\end{align}
In general, for two unaries $u_1$ and $u_2$
$u_1 \sqcup u_2$ is a unary logical form representing:
\begin{align}
\tolc{u_1 \sqcup u_2} = \lambda x . \tolc{u_1}(x) \vee \tolc{u_2}(x)
\end{align}

\paragraph{Negation}

Negation is straightforward.  Here is \nl{states not bordering California}
in lambda DCS:
\begin{align}
\wl{Type}.\wl{USState} \sqcap \neg \wl{Border}.\wl{California}
\end{align}
and in lambda calculus:
\begin{align}
\lambda x . \wl{Type}(x, \wl{USState}) \wedge \neg \wl{Border}(x, \wl{California}).
\end{align}
In general, for a unary $u$,
\begin{align}
\tolc{\neg u} = \lambda x . \neg \tolc{u}(x).
\end{align}

\paragraph{Higher-order functions}

Higher-order functions operate on sets of entities rather than on individual
entities.  These include aggregation (counting, finding the min/max),
superlatives (taking the argmin/argmax),
and generalized quantification.  In lambda DCS, these are implemented in the
natural way.

For example, the number of states and the number largest state by area are represented as
follows in lambda DCS:
\begin{align}
  & \wl{count}(\wl{Type}.\wl{USState}), \\
  & \wl{argmax}(\wl{Type}.\wl{USState}, \wl{Area}).
\end{align}
In lambda calculus:
\begin{align}
  & \wl{count}(\lambda x . \wl{Type}(x, \wl{USState})), \\
  & \wl{argmax}(\lambda x . \wl{Type}(x, \wl{USState}), \lambda x . \lambda y . \wl{Area}(x, y)).
\end{align}
In general, for an aggregation operator $A$ (e.g., $\wl{count}$) and a unary $u$, we have:
\begin{align}
\tolc{A(u)} = \lambda x . [x = A(\tolc{u})].
\end{align}
For a superlative operator $S$ (e.g., $\wl{argmax}$), unary $u$, and binary $b$:
\begin{align}
\tolc{S(u, b)} = \lambda x . [x \in S(\tolc{u}, \tolc{b})].
\end{align}

\paragraph{Lambda abstraction}

Up until now, our lambda DCS logical forms have been limited in two ways:
\begin{enumerate}
\item Fundamentally, our logical forms have been
tree-structured, since the only way different parts of the logical forms can
currently interact is via one implicitly represented overlapping variable.
Non-tree-structured logical forms are important for handling bound anaphora,
and to capture such phenomena, we will introduce a construct called \emph{mu abstraction}.

\item All our composition operations have created unaries.
However, higher-order operations such as $\wl{argmax}$ take in a binary which
supply the comparison function, which could be non-atomic.
To construct binaries compositionally, we introduce lambda abstraction,
which is in general distinct from lambda abstraction in lambda calculus,
although the two coincide in some cases.
\end{enumerate}

% mu
Let us start with an example of bound anaphora:
\nl{those who had a child who influenced them}.
In lambda DCS, we use mu abstraction:
\begin{align}
  \mu x . \wl{Children} . \wl{Influenced} . x
\end{align}
Here, the $\mu x$ simply adds a constraint that the first argument of $\wl{Children}$
be bound to the second argument of $\wl{Influenced}$.
In lambda calculus, this expression is:
\begin{align}
\lambda x . \exists y . \wl{Children}(x, y) . \wl{Influenced}(y, x).
\end{align}

% lambda
Let us consider a superlative with a compositional comparison function:
\nl{person who has the most children}.
In lambda DCS, we use lambda abstraction to construct a new binary logical form
denoting pairs of people and the number of children they have:
\begin{align}
  \wl{argmax}(\wl{Type}.\wl{Person}, \reverse{\lambda x . \wl{count}(\reverse{\wl{Children}}.x})),
\end{align}
where we use the \emph{reverse operator} $\tolc{\reverse{b}} = \lambda y . \lambda x . \tolc{b}(x, y)$
switches the arguments of $b$.

The equivalent logical form in lambda calculus:
\begin{align}
\wl{argmax}(\lambda x . \wl{Type}(x, \wl{Person}),
            \lambda x . \wl{count}(\lambda z . \wl{Children}(x, z))).
\end{align}

In general, for a variable $a \in V$ and a unary $u$, we define the
conversion for variables, mu abstraction, and lambda abstraction as follows:
\begin{align}
\tolc{a} &= \lambda x . [x = a], \\
\tolc{\mu a . u} & = \lambda x . [a = x] \wedge \tolc{u}(x), \\
\tolc{\lambda a . u} &= \lambda x . \lambda a . \tolc{u}(x).
\end{align}
Note that applied to unaries, lambda abstraction produces binary logical forms,
while mu abstraction produces unary logical forms.

\subsection{Discussion}

A prevailing theme in lambda DCS
is that the construction of logical forms
centers around \emph{sets} (of entities or entity pairs), whereas in lambda
calculus, construction centers around truth values.  Lifting up to sets allows
us to eliminate variables, treating them as implicitly existentially
quantified.  Indeed, looking back to the conversion function $\tolc{\cdot}$,
they all have the form $\tolc{z} = \lambda x . \text{something}$.
Of course, there are logical forms where lambda DCS would not offer much savings
in reducing variables---for example, when many predicates operate on the same
variables in a non-tree-structured manner.  However, we have found that for logical
forms derived from natural language, especially for querying databases,
lambda DCS is a good choice.

% Connection to other paradigms, DCS
The development of lambda DCS has been inspired by many other formalisms.
As evident by name,
lambda DCS is adapted from Dependency-Based Compositional Semantics (DCS)
\citep{liang11dcs}, which prominently features (i) tree-structured logical forms and
(ii) default existential quantification, the latter being imported from Discourse
Representation Theory \citep{kamp93drt}.
There are two major differences between lambda DCS and DCS.
First, lambda DCS revolves around unaries and binaries, which we found to be
the right level of generality; DCS allowed arbitrary arities,
and thus was a little too low-level.
Second, we have omitted the mark-execute construct of DCS, which is an
orthogonal direction for future exploration.  We can think of mark-execute
as more in the realm of constructing logical forms compositionally rather than
their representation.  We have focused on the latter here.

% Description logic
The syntax of lambda DCS and the tree-structured nature (without variables) is
derived from the concept constructors in description logic \citep{baader2003description},
but the marriage with variables and higher-order functions is specific to
lambda DCS.  Of course, description logic is designed for logical inference,
which necessitates a simpler logic, whereas lambda DCS is designed for model
checking (e.g., querying a database), and can therefore bear the
additional complexity.

% SPARQL
Given our application to querying graph databases, it is natural that there are
some parallels between lambda DCS and graph query languages such as SPARQL
\citep{harris2011sparql}, to which we ultimately convert in order to execute
the logical forms \citep{berant2013freebase}.
Working with a graph database leads to thinking about logical forms as graph
patterns, although this intuition is less helpful when we add higher-order
operations.

% Expressiveness
In terms of expressivity, lambda DCS is clearly dominated by lambda
calculus, as evident by our conversion from the former to the latter.  From
this perspective, lambda DCS is syntactic sugar.  However, we believe that
lambda DCS is more closely connected with the compositional structure of
natural language and therefore can be quite convenient to work with.

% Construction
We should also note that lambda calculus in formal semantics is really used for
two purposes, which we'd like to dissect:
The first purpose is to construct the logical form of a sentence compositionally,
where lambda abstraction is merely used as a macro to piece bits of logical form
together.
The second purpose is to represent the final logical form,
where lambda abstraction is needed to denote a function
(e.g., the comparison function passed into $\wl{argmax}$).
Here, we have mainly focused on the latter representation issues.
As for construction, we have so far
used a very simple construction mechanism
\citet{berant2013freebase},
since we targeted the semantic parsing of short
questions, which have limited compositional demands.
In principle, lambda DCS logical forms could be constructed using lambda
calculus (serving as a macro) in a CCG formalism.
Analogously, lambda calculus has been used to build logical forms in Discourse
Representation Theory \citep{muskens96combine}.
Down the road, we hope that working with lambda DCS can lead to new
construction mechanisms suitable for complex sentences.

\bibliographystyle{plainnat}
\bibliography{all}

\end{document}